\documentclass[letter,10pt,conference]{ieeeconf}
\IEEEoverridecommandlockouts
\usepackage{amsmath,amssymb,theorem}
\usepackage{graphicx}
\graphicspath{ {./images/} }
\usepackage{algorithm}
\usepackage{algpseudocode}
\usepackage{psfrag}
\usepackage{cite}
\usepackage{rotating}

\usepackage{color,xspace}
\usepackage{subfigure}
\usepackage{microtype}
\newtheorem{theorem}{Theorem}[section]

\newtheorem{problem}[theorem]{Problem}

\newcommand{\real}{{\mathbb{R}}}

\newcommand{\PP}{{\mathcal{P}}}
\newcommand{\UU}{{\mathcal{U}}}

\newcommand{\timestep}{\Delta t}

\newcommand{\oprocendsymbol}{\hbox{$\bullet$}}
\newcommand{\oprocend}{\relax\ifmmode\else\unskip\hfill\fi\oprocendsymbol}

\title{\LARGE \bf Agent-Based Emulation for Deploying Robot Swarm Behaviors}

\author{Ricardo Vega\qquad Kevin Zhu \qquad Connor Mattson \qquad Daniel S. Brown \qquad Cameron Nowzari \thanks{R. Vega, K. Zhu, and C. Nowzari are
    with the Department of Electrical and Computer Engineering,
    George Mason University, Fairfax, VA 22030, USA, {\tt\small
      \{rvega7, kzhu4, cnowzari\}@gmu.edu}}
        \thanks{
      C. Mattson and D. S. Brown are
    with the Kahlert School of Computing,
    University of Utah, Salt Lake City, UT 84112,  USA, {\tt\small \{c.mattson,daniel.s.brown\}@utah.edu}}}

\begin{document}
\maketitle 

\begin{abstract}

Despite significant research, robotic swarms have yet to be useful in solving real-world problems, largely due to the difficulty of creating and controlling swarming behaviors in multi-agent systems. Traditional top-down approaches in which a desired emergent behavior is produced often require complex, resource-heavy robots, limiting their practicality. This paper introduces a bottom-up approach by employing an Embodied Agent-Based Modeling and Simulation approach, emphasizing the use of simple robots and identifying conditions that naturally lead to \textit{self-organized} collective behaviors. Using the Reality-to-Simulation-to-Reality for Swarms (RSRS) process, we tightly integrate real-world experiments with simulations to reproduce known swarm behaviors as well as discovering a novel emergent behavior \textit{without} aiming to eliminate or even reduce the sim2real gap. This paper presents the development of an Agent-Based Embodiment and Emulation process that balances the importance of running physical swarming experiments and the prohibitively time-consuming process of even setting up and running a single experiment with 20+ robots by leveraging low-fidelity lightweight simulations to enable hypothesis-formation to guide physical experiments. We demonstrate the usefulness of our methods by emulating two known behaviors from the literature and show a third behavior `discovered' by accident.  
\end{abstract}

\section{Introduction}\label{se:intro}

Swarms are purported to be useful in many real world applications including pollution monitoring \cite{GZ-GKF-DPG:11}, disaster management systems \cite{HK-CWF-IT-BS-ER-KP-AW-JW-12}, surveillance \cite{MS-JC-LP-JT-GL-AT-VV-VK:14}, and search and rescue \cite{RA-JJ-BA-EM:20}; but after 50+ years of research we still do not see any engineered swarms solving practical problems or providing any real benefits to society. This may be due to the fact that we are yet to fully understand how we can control swarms of robots due to the naturally complex/chaotic interplay between agents that leads to emergent behaviors~\cite{JT:05, JT:05ten, OTH:07}. This makes it difficult to reproduce emergent behaviors found in virtual environments in physical ones~\cite{HH-TA-AR:20, GV-CV-GS-TN-AEE-TV:18}. 

The traditional approach to producing robot swarms has been to use top-down methods with a specific macro objective or metric in mind that the multi-agent system should optimize~\cite{AO-MG-AK-MDH-RG:19, MG-JC-WL-TJD-RG:14, DS-CP-GB:18, MG-JC-WL-TJD-RG:14clust}. This results in the engineering of carefully controlled agent interactions to guarantee certain macro-behaviors `emerge.' This generally requires agents to be able to locally monitor something and regulate it (e.g., distance to the nearest neighbor). 
Though this method may eventually result in working multi-robot systems, it also frequently requires expensive and highly-capable robots to coordinate with each other using complex control algorithms and generally uses a lot of resources and time to design.  For example, there are many collective swarm behaviors that have been well studied including flocking, shepherding, shape formation, and collective transport; however, the agents used to create those behaviors are rather sophisticated and often require sensing/actuating/communicating capabilities that are not practically scalable in the real world~\cite{ALA-MGCAC-NDF-ML-GV:18, YC-YRH-MRD-ALB:07, AO-MG-RG:17, YM-HG-YJ:13, MR-AC-JW-GH-JM-RN:13}.

In contrast, the key aspect of natural biological swarms is true emergence where the distance between agents is not directly controlled, but \textit{is} the emergent property of more internal biological autonomic reactions. In this project we similarly only study simple agents with direct sensor-to-action (non-symbolic) mappings rather than the more traditional sense-\textit{plan}-act framework for robotics~\cite{HVDP-SB-JO:03, ALA-MGCAC-NDF-AL-ML-GV:18}. 
By finding methods of producing various emergent behaviors in physically embodied~\cite{RAB:90, RAB:91} experiments that don't require any perception/computation, we offer the opportunity the immediately scale up any physical swarm system by removing the most expensive part of traditional robots not just financially but also through setup/programming/debugging time -- the CPU.

\paragraph*{Statement of Contributions}
This paper presents an alternate approach to the prohibitively large sim2real gap in swarm systems due to the inherently complex/chaotic dynamics of such systems. Instead of aiming to transfer anything found in simulation to reality directly, we acknowledge both exploration and validation of robot swarms will need to happen in physically embodied agents. Monte-Carlo physical experiments are impractical so we guide our experiments with low-fidelity simulations. This paper presents an experimental swarm discovery methodology. 

First, we show how we build a low-fidelity simulator with low model confidence but sufficient value to guide physical experiments. The simulator is based on experimentally measured data from individual robot characteristics, including their idiosyncrasies (small agent-to-agent differences). Second, we use phase diagrams created from simulations to allow the formation of hypotheses that can be tested on the physical robots. 
Finally, we demonstrate the potential of non-symbolic controllers to exhibit higher levels of complexity through inter-agent interactions. All results are validated on physical experiments.

\section{Relevant Literature}
Although there has been much work focusing on `swarm' systems, the distinction between a swarm and a multi-agent system isn't always clearly identified. Many may consider these to be one in the same, however here we suggest that this isn't always the case. There is clearly a difference between the composition of swarms in nature (e.g., murmurations of starlings, bait balls of sardines, or ant mills) and a few drones  being instructed to fly together. Not only is there a difference in the number of individuals, but, more importantly, there is also the aspect of emergence, specifically \textit{self-organization}, in a swarm that isn't always necessarily present in a multi-agent system.

\paragraph*{Swarms and Emergence}
The common general definition of emergence within a system is when local interaction between agents at a lower-level alters the higher-level properties of the system. Multiple different works \cite{JT:05, JT:05ten, OTH:07} have identified and defined different types of emergence; these `types' of emergence are identified depending on not only how and to what extent the lower level of the system affects the higher level, but also on the top-down feedback (i.e. how changes in the higher level in turn affect the low level constituents) ranging from Type I intentional simple emergence (a clock made up of individual components) to Type IV strong emergence (genetic evolution). Self-organization, or the process by which individuals create global behaviors through just the interactions amongst themselves rather than through external intervention or instruction \cite{DW:06}, can occur in any type of emergence (besides intentional simple emergence). This paper defines a swarm to be a multi-agent system that exhibits self-organization at some level and focuses only on the simplest type of emergence where self-organization can occur: \textbf{unintentional simple emergence}, defined as the properties/behaviors that arise as a result of agent interaction that are only applicable to the group. This is in contrast to the myriad of swarming works that study \textbf{weak emergence} in which some local quantity like distance or alignment is directly considered and controlled using a symbolic or even sub-symbolic approach~\cite{CT-CL-CN:20, UB-LW-ADA-ME:15, KS-IBS-LMTR-CRH-DM-MAH:16}.

\paragraph*{Swarm Chemistry}

Our bottom-up approach to discovering and analyzing swarm behaviors is heavily influenced by the field of chemistry. Connections between swarms and chemistry are not entirely new: the idea of `swarm chemistry'  was first coined by Sayama in 2009~\cite{HS:09} where it was found that mixing virtual agents with different control rules led to fascinating new behaviors~\cite{HS:09, HS:11, HS:12}. Due to the very simply parameterized simulated agents, new swarm `recipes' that lead to new complex behaviors are discovered regularly; even by undergraduate students~\cite{HS:07}. Our goal is exactly to replicate this model in physical swarming experiments. The three major challenges we have to confront in the real world are (i) idiosyncrasies of the robots, (ii) physics, and (iii) the inability to run experiments at scale. While the agent-based simulation community enjoys Monte-Carlo data collection with noise-free data and fully repeatable experiments, we can only collect noisy data from non-repeatable experiments at a rate many orders of magnitude slower.

\paragraph*{Agent-Based Simulation to Reality (sim2real) Gap}
Simulations are a very natural and effective way of addressing the three challenges above and works very well for single plant/robot systems~\cite{VL-HH-LYC-JW-JI-DS-ML-KG:21, YC-AH-VM-MM-JI-NR-DF:19, LW-RG-QV-YQ-HS-HC:22}. For swarms, although agent-based modeling and simulations are the de-facto standard in studying emergent behaviors \cite{OP-GE-AJC:09, VD:94, JT:05ten, YBY:19, RC-CC:96, BE:05, JHH:00}, the vast majority of these works start and end in strictly virtual environments. Additionally, there are various theoretical conditions and even guarantees that have been provided for swarming behaviors~\cite{HS-HKG:22,ZC-ZC-VT-DC-HZ:16,MRD-YC-ALB-LSC:06, NVB-HA-IYT-SAM:20}, and although they may be useful to some extent, it is important to note that often these claims only consider the model that represents the real robots and therefore only truly hold in theory/simulation and, due to the reality gap, shouldn't be relied on to always hold with real robot systems. 

\begin{figure}[h!]
   \centering
   \includegraphics[width=.75\linewidth]{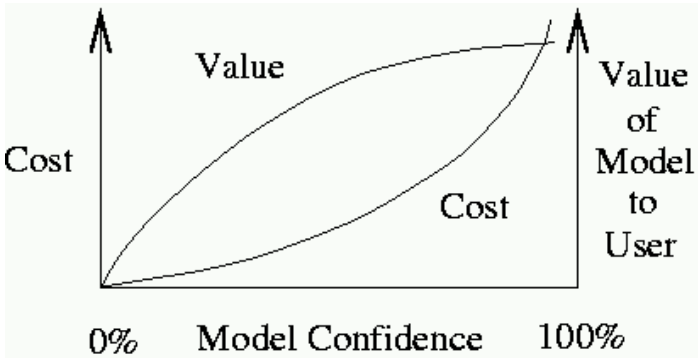}
   \caption{Diminishing return of model value relative to development cost~\cite{RGS:10}.}
   \label{fig:modelusefulness}
\end{figure}

Due to the inherently chaotic trajectories of swarm systems, we know that even with a lot of resources spent on developing a simulator, it will still not yield a sufficiently high model confidence or value to enable direct simulation-to-reality transfers. Consequently, we aim to instead get 80\% of the value of simulation by spending only 20\% of the time/cost that may normally be spent in simulation ahead of physical experiments~\cite{RGS:10}; see Figure~\ref{fig:modelusefulness}. This allows the human researchers to still gain valuable insights into what-if scenarios of swarms without being overly reliant on refining it in simulation only to find it doesn't work the same way in real environments.

\newcommand{\omegamax}{\omega_\text{max}}

\section{Problem Formulation}\label{se:problem_formulation}

The overarching questions we aim to answer are simply how can we rapidly (i) deploy limited-capability robots to produce predictably emergent behaviors, and (ii) discover new complex behaviors from simple interactions?

The easy answer would be to skip simulation altogether and if time/money were not factors, running millions of experiments and carefully recording observations as a biologist/scientist would prove invaluable to better understanding embodied self-organizing behaviors. Since Monte-Carlo physical experimentation of large numbers of robots is not pragmatic, we employ a lightweight simulator-in-the-experimentation-loop approach to guide specifically which physical experiments should be prioritized. We will first show how already-known emergent behaviors using non-symbolic controllers can be quickly reproduced on our platform using a low-fidelity NetLogo simulation.

The starting point of our problem is the availability of a team of~$N$ homogeneous but idiosyncratic robots (i.e., slightly different from robot to robot) with general unknown dynamics
\begin{align}\label{eq:realdynamics}
    \dot{x}_i = f(x_i,u_i,\theta^a_i,w_i),
\end{align}
where~$f : \real^d \times \UU \times \real^p \times \real^d \rightarrow \real^d$ is an unknown function of a robot's state~$x_i \in \real^d$, its control input~$u_i \in \UU \subset \real^m$, the actuation idiosyncrasies~$\theta^a_i \in \real^p$, and the unmodeled disturbances~$w_i \in \real^d$. Depending on the physical sensor(s) used, the ultimate goal is to directly connect these outputs to drive actuators when ready to scale up essentially a swarm of Braitenberg vehicles~\cite{VB:96}. As this paper is focused on the exploration and discovery of finding useful hard-wireable reactive controllers, our robots do use a CPU and software to parameterize the map from sensor readings to actuators. 

\subsection{Reproducing known behaviors}

Unfortunately, given this starting point there are no existing works that can help guide exactly how to realize a known collective behavior on the real robots, let alone a novel behavior. While many different groups have shown the success in producing group behaviors with their robots, the process of replicating the results on a different set of robots isn't necessarily straightforward. 

With the basic binary sensing capability set given to these robots and as long as the robots have dynamics~\eqref{eq:realdynamics} that support unicycle-like kinematics with a forward speed~$v$ and turning rate~$\omega$, it is known that some various emergent behaviors are possible such as \textbf{milling}, where the agents form a rotating circle; and \textbf{diffusing}, where agents spread out evenly over the environment~\cite{MG-JC-TJD-RG:14, DB-RT-OH-SL:18, DS-CP-GB:18, FB-MG-RN:21}. These \textit{simply} emergent behaviors only appear under certain nontrivial conditions that have not yet been fully characterized (especially milling), thus we will investigate conditions that yield higher chances of producing these behaviors with our robots. This is in contrast with \textit{weakly} emergent behaviors that can produce the same effect but rely on a higher complexity local controller such as monitoring/controlling local distances or even computing centroids of complicated Voronoi cells~\cite{CN-JC:11-auto,JC-SM-TK-FB:02-tra}. 

Our starting point is the simple binary sensing-to-action controller that turns left when it detects anything, and turns right otherwise. This controller enables \textbf{milling}~\cite{FB-MG-RN:21, DS-CP-GB:18},

\begin{equation} \label{eq:mill_controller}
\begin{split}
u_{i}^\text{milling}(\ell) &= \begin{cases} (v,\omega), \quad &\text{if } y_i = 1 , \\ (v,-\omega), \quad &\text{otherwise.} \end{cases} 
\end{split}
\end{equation}
Instead, by moving backwards if anything is detected or otherwise turning in place, the controller can lead to \textbf{diffusing}~\cite{AO-MG-AK-MDH-RG:19},

\begin{equation} \label{eq:diff_controller}
\begin{split}
u_{i}^\text{diffusing}(\ell) &= \begin{cases} (-v,0) \quad &\text{if } y_i = 1 , \\ (0,\omega) \quad &\text{otherwise.} \end{cases}  
\end{split}
\end{equation}

\begin{figure}[t]
    \centering
    \includegraphics[width=1\linewidth]{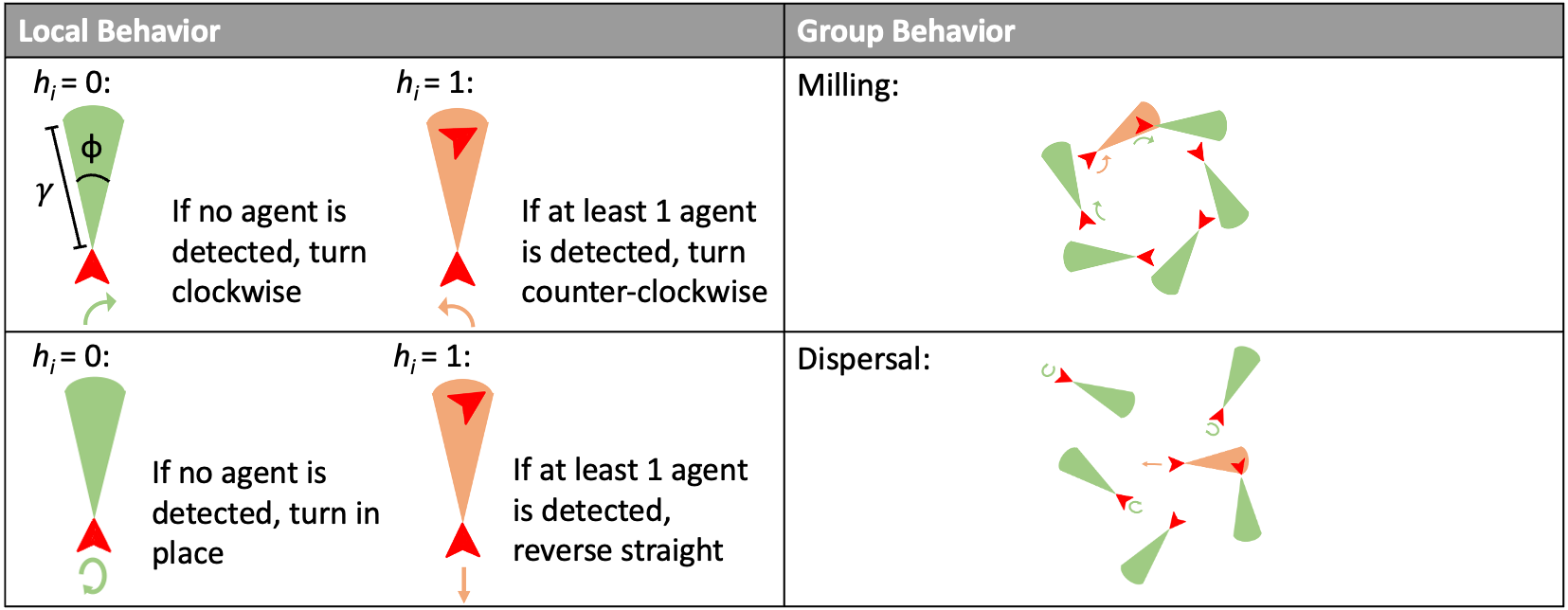}
    \caption{Different local interaction rules leading to different emergent group behaviors (under certain conditions).}
    \label{fig:milling_figure}
\end{figure}

Figure~\ref{fig:milling_figure} shows these simple direct sensor-to-action controllers and the group behavior they have the potential to produce. 
Because our robots are not able to plan, we must instead understand the conditions which enable the behaviors.

\begin{problem}[Real-world Swarm Behaviors]\label{pr:main}
{\rm
    In order to deploy a real world robot swarm experiment one is quickly faced with a lot of unanswered questions:
    \begin{enumerate}
        \item How many robots should be deployed?
        \item How fast should they be moving?
        \item How good do the sensors need to be?
        \item How homogeneous does the swarm need to be?
    \end{enumerate}
}
\end{problem}

Problem~\ref{pr:main} is not an exhaustive list of questions that need answers and is completely ignoring environmental conditions for instance. The existing works that produced these behaviors so far required a lot of trial and error in physical experiments due to a lack of a way to do standard hypothesis testing. We aim to use simulations in a very fast way that still gives us a nonzero chance that our very first experiment might actually succeed rather than starting a search completely in the dark. 

For instance without any simulation data, when a swarming behavior fails it may be inadvertently assumed that more robots are needed but this paper provides evidence it is not that simple. Instead, it reveals how complicated searching this space for a successfully emergent behavior can be. If the first experiment deployed doesn't work, how does one know what to modify for the second experiment? Did the experiment fail because the sensors weren't good enough, they were moving too fast, or maybe there were \textit{too many} robots on the field? To answers these questions we develop phase diagrams~\cite{AC-CKH:18, MRD-YC-ALB-LSC:06, NVB-HA-IYT-SAM:20, ZC-ZC-VT-DC-HZ:16} to gain insight into our embodied problem. The difference here is the use of simulations-in-the-design-loop to more effectively navigate the physical space and better understand why experiments fail when they do. This enables a much more organized and systematic approach to deploying and testing robot swarm systems than is done today.

\subsection{Discovering new behaviors}
While validating and reproducing behaviors found in other research groups is fine, our true aim is to make this process of discovering simple mechanisms leading to complex behaviors in the real world more efficient and useful. 
Similar to how colonies of ants or bees have a well understood collective intelligence, we wish to understand how mixing different species of physically embodied agents can self-organize into more complex structures. With the same binary sensing capability, here we will consider the exploration of mixing two different modes of reactive controllers in search for new emergent behaviors. For now this is a human-intensive task but fortunately is one that can be delegated to undergraduate or even high school students. 

\begin{figure*}[t]
\centering
    \includegraphics[width=.85\linewidth]{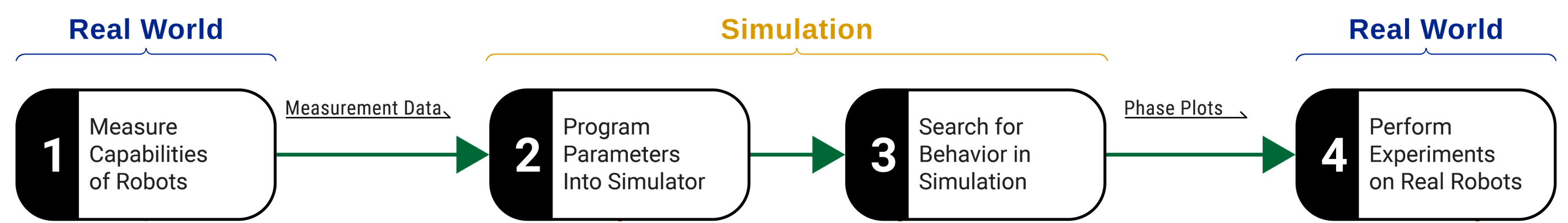}
    \caption{Flow Chart of Reality-to-Simulation-to-Reality for Swarms (RSRS) Process}\label{fig:steps}
\end{figure*}

\section{Agent-Based Modeling and Simulation}\label{se:rsrsr}
We aim to make use of the `Reality-to-Simulation-to-Reality for Swarms' (RSRS) process shown in Fig.~\ref{fig:steps} to methodologically and quickly answer Problem~\ref{pr:main} in four main steps

\begin{enumerate}
    \item Measure the capabilities/dynamics of robots;
    \item Build simulator;
    \item Explore within simulation;
    \item Run real robot experiment based on phase diagrams.
\end{enumerate}

First, the individual robots’ capabilities are characterized and experimentally measured. Second, these capabilities are abstracted into a simple simulator that aims to gain value without large modeling cost (cf. Fig.~\ref{fig:modelusefulness}).
Third, equipped with a simple agent-based simulator, large amounts of data can be collected to gain insights and form hypotheses about physical experiments. Finally, the discoveries found in simulation need to be evaluated using the real set of robots to be practical.

The details on the first two steps are in the Appendix. In Section~\ref{se:case} we show the final step of performing physical experiments guided by phase diagrams created in Step 3. 
The remainder of this section discusses the details of Step 3 and how we know when we have found something.

The controllers~$u^\text{milling}$~\eqref{eq:mill_controller} and $u^\text{diffusing}$~\eqref{eq:diff_controller} are too simple to always guarantee milling and diffusing, respectively. To study the conditions under which they do, we denote $\PP$ as the parameter space~$\real_{+}^3 \times [0, 2\pi) \times \mathbb{N}$, containing the agent parameters of vision-distance $\gamma$, forward speed $v$, turning rate $\omega$, vision-cone $\phi$ and the total number of agents $N$.

\begin{figure}[t]
\centering
  \includegraphics[width=0.98\linewidth]{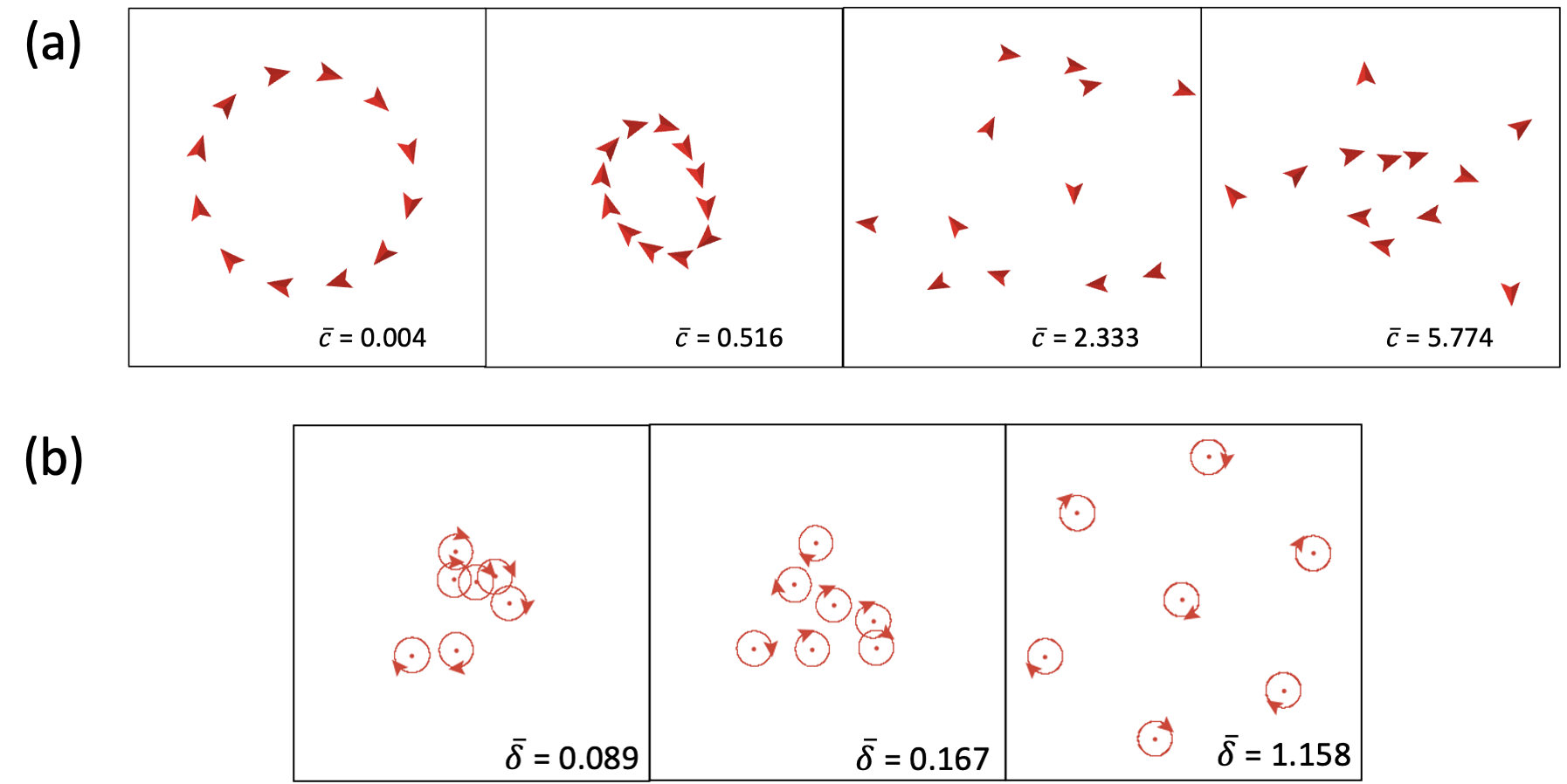}
  \caption{Examples of different values of (a) $\overline{c}$ and (b) $\overline{\delta}$.}
  \label{fig:metrics_plots}
\end{figure}

The key first part of this exploration step is to identify or define the macroscopic metrics that should be used to differentiate and recognize when the sought after behavior occurs. In regards to the milling behavior, we chose to measure the `circliness' of the system, similar to the metrics in \cite{CT-CL-CN:20}, which considers the distance of the closest and furthest agent from the center of mass $\mu = \frac{1}{N}\sum^{N}_{i=1}\mathbf{p}_i(t)$ with

\begin{equation}\label{eq:circliness}
\overline{c} = \frac{\max_{i \in N} ||\mathbf{p}_i-\mu|| - \min_{i \in N}||\mathbf{p}_i-\mu||}{\min_{i \in N}||\mathbf{p}_i-\mu||},
\end{equation}
where~$\overline{c}=0$ represents a perfect mill. Fig~\ref{fig:metrics_plots}(a) shows snapshots of what different values of~$\overline{c}$ look like.  Similarly, we include a metric that measures how well the swarming is diffusing with
\begin{equation}
    \overline{\delta} = \frac{\min_{i,j \in N}||\mathbf{o}_i - \mathbf{o}_j||}{\gamma},
\end{equation}
where $\overline{\delta} > 1$ represents a diffused system. This latter metric involved calculating the center of the circular path of the agents (i.e. the pivot), $\mathbf{o}_i$, and comparing the minimum distance between these points and their vision distance. With this metric, we can better identify the diffusion behavior from any dispersal behavior that would be considered the same using something like a scatter metric (i.e. average distance from center of mass).
  \begin{equation} \label{eq:pivot}
\begin{split}
o_{i,1}(\ell) &= z_{i,1}(\ell) + \frac{u_{i,1}(\ell)}{u_{i,2}(\ell)} sin(z_{i,3})(\ell)  \\
o_{i,2}(\ell) &= z_{i,2}(\ell) + \frac{u_{i,1}(\ell)}{u_{i,2}(\ell)} cos(z_{i,3})(\ell)  
\end{split}
\end{equation}

\begin{figure}[h]
    \centering
   
     	{\includegraphics[width=.95\linewidth]{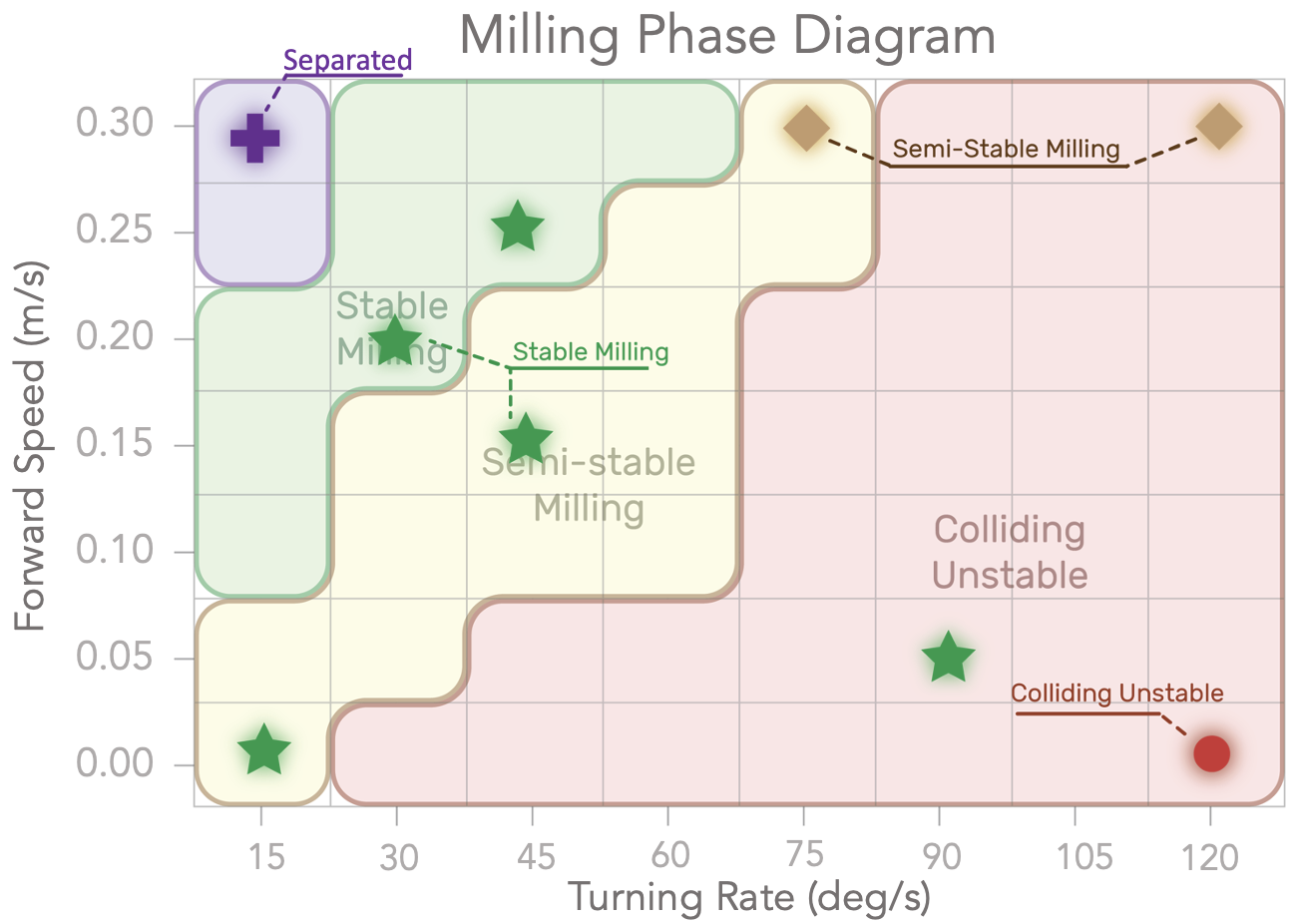}}
    \caption{A 2D slice of a phase diagram produced using simulated results for the milling behavior demonstrating the value of even a low-fidelity phase diagram as a guide to physical experiments. The 9 colored shapes in the phase diagram show that only 5 of 9 physical experiments matched the simulator results, but the usefulness in the insights gained from the 56 simulations with only 50\% model confidence still has clear value to humans and enables forming hypotheses that can be physically tested.}
    \label{fig:phase_plot2}
\end{figure}

To create the phase diagrams, we consider the parameters $\PP$ to be the independent variables that can be changed (e.g. $N, v, \omega$). We also simulate cases where the values may be outside the limits given the current sensors and actuators on the robots so that we may be aware of what upgrades should be done to get to behaviors not possible at the present state. Given the size of the phase space, multiple different phase diagrams can be created for each behavior. An example of a phase diagram that was created for the milling behavior can be seen in Fig.~\ref{fig:phase_plot2}, where various values of $\omega$ and $v$ were simulated. The green region represents when the agents in the simulator produced a stable well-shaped circle (i.e. $\overline{c} < 0.2$) and the yellow area represents simulations where circliness was measured to be between $0.2$ and $1$ (ellipsoidal rather than circular). The sims in the red and purple regions both had circliness values over $1$, though they resulted in either colliding clusters or separated groups, respectively.

With all this information, we now have a better idea of how the robot experiment should be setup to give us a better chance of creating the milling and the diffusing behaviors, and also gives us better insight how we should reprogram and redeploy the robots if they create an unwanted behavior.

\section{Agent-Based Embodiment and Emulation}\label{se:case}
Here we show how we can work towards Problem~\ref{pr:main} and validate two known behaviors on real robots by applying the RSRS framework discussed in Section~\ref{se:rsrsr} to the TurboPi robots as well as discuss how we discovered a unexpected group behavior.

\subsection{Reproducing known behaviors}
\begin{figure}[t]
    \centering
    \includegraphics[width=0.98\linewidth]{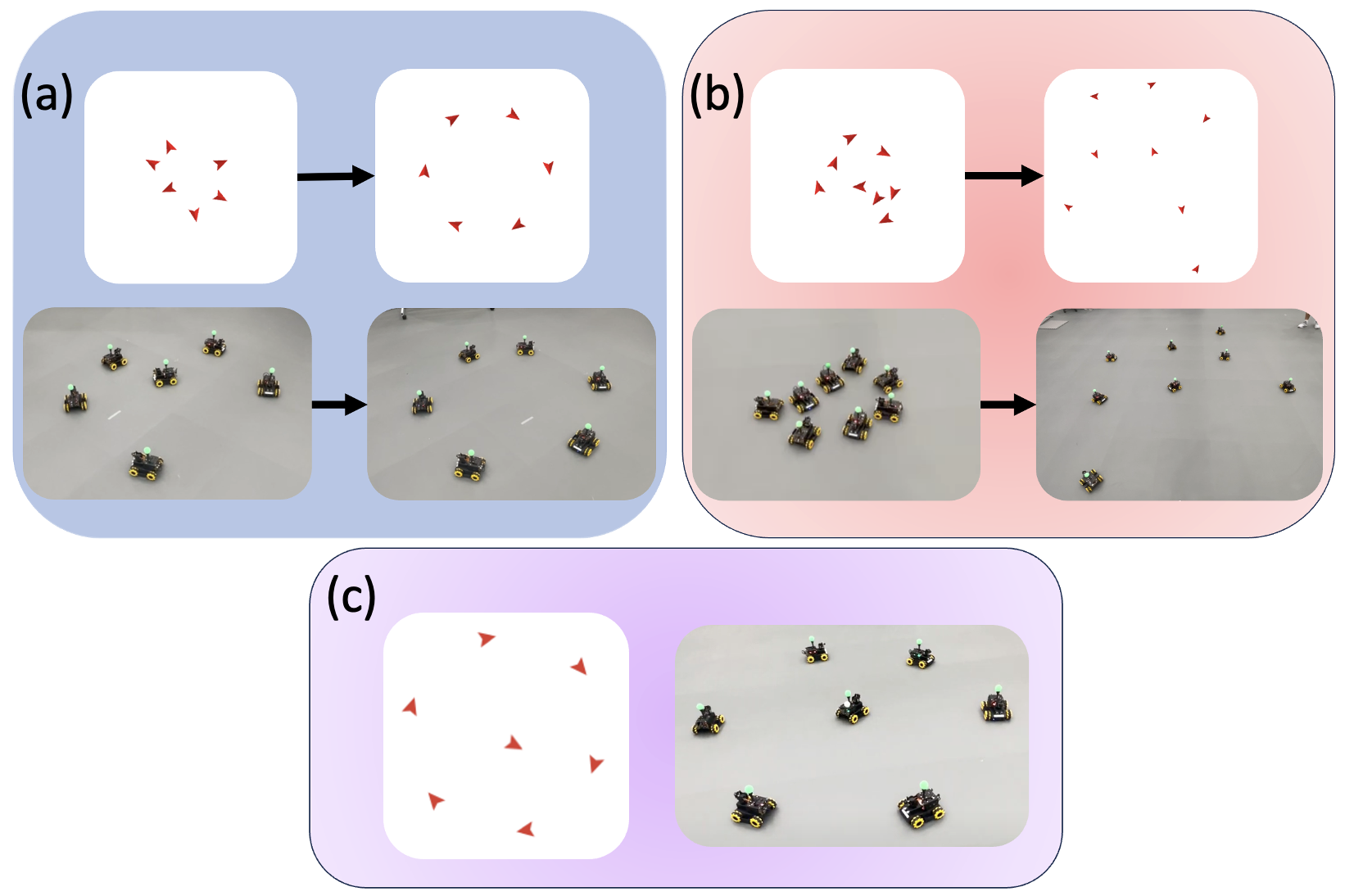}
    \caption{(a) Start and finish of six agents milling in NetLogo simulator and six TurboPi successfully milling using the controller~\eqref{eq:mill_controller} with $v = 0.25\frac{m}{s}$ and $\omega = 45 \frac{deg}{s}$. (b) Start and finish of eight agents diffusing in NetLogo simulator eight TurboPi successfully diffusing using the controller~\eqref{eq:diff_controller} with $v = 0.3\frac{m}{s}$ and $\omega = 150 \frac{deg}{s}$. (c) Unexpected self-centering behavior discovered through RSRS process.}
    \label{fig:sim-real-experiments}
\end{figure}

With the information provided from Step 3 of the RSRS process, we can begin running experiments on the robots using the recommendations that, according to our simulations, has the highest chance of producing the desired diffusing and milling behavior. Remarkably, both behaviors were successfully recreated with the real TurboPis given the parameters from the phase diagram within the first few experimental attempts. Both the start and the finish of the simulated and real experimental trials are shown for milling in Fig.~\ref{fig:sim-real-experiments}(a) and for diffusion in Fig.~\ref{fig:sim-real-experiments}(b).

\subsection{Discovering new behaviors}
Through this process, we were able to successfully reproduce known behaviors on an existing system of real robots. On top of this, while running milling experiments on the real robots, we observed a faulty robot continuously missing the robot in front of it and moving inward, causing the rest of the mill to break apart and have to reform. Upon seeing this, we went back to the simulator and explored what could happen if one of the robots was running a different controller. Through this second phase of exploration, we discovered that because of the actuating and sensing noise, one agent running the controller:
\begin{equation} \label{eq:self_center_controller}
\begin{split}
u_{i}^
\text{self-centering}(\ell)&= \begin{cases} (v,\omega), \quad &\text{if } y_i = 1 , \\ (0,-3\omega), \quad &\text{otherwise.} \end{cases} 
\end{split}
\end{equation}
where $v$ and $\omega$ are the same used by the rest of the agents running controller~\eqref{eq:mill_controller}, would lead to a self-centering behavior that causes the system as a whole to produce a `Bulls-eye' shape. This additional behavior would have gone unnoticed/undiscovered if we hadn't been running this iterative experimentation process and accounting for the imperfect idiosyncratic capabilities of the real robots.

\section{Conclusions}\label{se:conclusions}
In this paper, we hope to begin the proliferation of deploying swarms of robots in real environments to self-organize into useful behaviors. By using a low-fidelity simulator-in-the-design loop methodology we use large amounts of simulated data only to gain insight into how to prioritize which experiments should be run on real robots. More specifically, we show how even simple simulators can be used to help guide multi-robot control designs for homogeneous (but idiosyncratic) systems of robots. Not only were we able to validate the milling and diffusing behaviors, but through this process we discovered a new emergent behavior on real robots by mixing two different types of controllers. As evidenced by decades worth of research in discovering different swarming/emergent behaviors that have only been demonstrated in simulation~(e.g., Artificial Life, Swarm Chemistry),
we aim to replicate these discoveries on real robots that can then be immediately used in real environments.

\appendix \label{appendix}

 The TurboPi is a 19x16x14cm robot controlled via Raspberry Pi 4 and has four Mecanum wheels that give it omni-directional locomotive capabilities as well as two servo motors that allow the onboard camera to pan and tilt giving it a controllable larger Field-of-View (FOV). Additional sensors include an ultrasonic range-finder and a 4 channel line tracker. Although fitted with various sensors and more complex computation capabilities, our aim is to study the potential of non-symbolic controllers so for our purposes we only use one sensor (RGB camera) that generates a binary output of 1 when anything green is detected or 0 otherwise. We also kept the agent dynamics simple and had the robots operate under a unicycle model where the robot could only move forward or backwards and/or rotate its heading.

The first step of the RSRS process is to measure the capabilities/dynamics of the robots. We begin by simplifying the dynamics of the TurboPi robot cars to unicycle kinematics
\begin{align}\label{eq:real_func}
    g  = \left[ \begin{array}{c} u_{i,1} \theta^a_{i,1} \cos z_{i,3} \\ u_{i,1} \theta^a_{i,1} \sin z_{i,3} \\ u_{i,2} \theta^a_{i,2} \end{array} \right].
\end{align}

In this case the reduced order model is given by~$n = 3$ states for each robot, with~$z_{i,1}$ and~$z_{i,2}$ representing the 2D position and~$z_{i,3}$ representing the orientation of robot~$i$.
The idiosyncrasies among the swarm are captured by~$\theta^a_i > 0$.

These robots used a controller with three inputs: speed, direction, and turning rate. However, since we wanted to simplify the robots to unicycle kinematics for now, we kept the direction of the robot to always be moving either forward in the direction the camera faced or in the opposite direction (i.e. reverse), therefore we can consider the robots to only have two inputs, speed and turning-rate. 

The range of these inputs were $[0,100]$ (though by changing the direction to be able to go in reverse, we can say it is actually $[-100,100]$) and $[-2,2]$ for the speed and turning-rate, respectively. These values are not in terms of any real units and therefore have no physical meaning. We don't necessarily need to make a full mapping of these values to real units at this point, simply measuring what the maximum and minimum of these ranges translate to in real life should be good enough to start.

To measure these actuation parameters, we used the OptiTrack system, a motion capture system that uses infrared cameras to track and analyze movements of objects with high-precision. Reflective markers were attached to the TurboPis which gave us the position and orientation of the robot as it moved across the environment at 120 frames per second, allowing us to calculate the speeds and turning rates.

 We performed multiple trials for various set of inputs across a representative group of five robots, a sample of these measurements for three different robots can be seen in Table~\ref{tab:speed1}. From these measurements, we found the mean and standard deviation for the measured speeds of all the robots, and for each robot, at each set value. The $\theta^a_1$ values of each tested robot was found by using the following equation:
\begin{align}\label{eq:theta^a1}
    \theta^a_1 = \frac{\text{Individual~Averaged~Speed}} {\text{Group~Averaged~Speed}}
\end{align}

We perform similar experiments and measurements to find the idiosyncrasies in turning rate~$\theta_2^a$ but omit the details here for brevity. 
By finding these distributions for the individual robots, we can predict the overall distribution of all measured robots to find how much the speeds and turning rate can vary in the system. Of course this data doesn't encompass all the information of the whole system but it gives us an idea of how inaccurate and idiosyncratic the robots may be. 
\begin{table}[h]
    \begin{center}
        \begin{tabular}{|l|l|l|l}
            \hline
            Inputs & Actual Speed Distribution ($\frac{cm}{s}$)            & $\theta^a_1$ \\
            \hline
            $u_1 = 50, u_2 = 0$           & $v_1$ $\sim$ $N(18.45, 0.06)$ & 1.00       \\
            $u_1 = 50, u_2 = 0$           & $v_2$ $\sim$ $N(18.90, 0.03)$ & 1.03       \\
            $u_1 = 50, u_2 = 0$           & $v_3$ $\sim$ $N(17.82, 0.05)$ & 0.97       \\
            $u_1 = 100, u_2 = 0$           & $v_1$ $\sim$ $N(28.17, 0.38)$ & 1.05       \\
            $u_1 = 100, u_2 = 0$           & $v_2$ $\sim$ $N(25.66, 0.12)$ & 0.95       \\
            $u_1 = 100, u_2 = 0$           & $v_3$ $\sim$ $N(27.04, 0.20)$ & 1.00       \\
            \hline
        \end{tabular}
    \end{center}
    \caption{Speed measurements of robots 1, 2, and 3 at various inputs}
    \label{tab:speed1}
\end{table}

Additionally, tests were also done to find the accuracy of the RGB camera sensors on board the robots. Rather than programming complicated code that recognizes other robots, we again kept it simple for now and simply attached a green colored ball. With this, we gathered data on the detection accuracy by moving the robot with the green ball attached to different distances and angles. It was found that, with the default sensor and color-detecting program, the detection region of the robots, on average, had a maximum distance of around 1.1 meters and an FOV of around 49 degrees. The robots' sampling period were measured at~$\timestep=22$ms or an operating frequency of~$\approx 45~$Hz.

With this information, we move towards the second step, building the simulator. For the sake of using an extremely intuitive and easy-to-modify simulator, we utilize NetLogo \cite{UW:99}, a multi-agent programmable modeling environment primarily used for educational purposes.

To program the idiosyncrasies, we first found the distribution of how far off the agents would move from the target parameter in the real world experiments and the reliability distribution of the RGB camera as mentioned in Step (1). Once we had a sample of the real distribution of noise, we expanded it such that the agents in the simulation would obtain their $\theta^a$ values from the normal distribution found from all the measured $\theta^a$ values.  

The discretized equations for the kinematic model are
\begin{align}\label{eq:disc_func}
g(\ell)  = \left[ \begin{array}{c} u_{i,1}(\ell) \theta^a_{i,1}(\ell) \cos z_{i,3}(\ell)\\ u_{i,1}(\ell) \theta^a_{i,1}(\ell) \sin z_{i,3}(\ell)\\ u_{i,2}(\ell) \theta^a_{i,2}(\ell)\end{array} \right], 
\end{align}
\begin{align}\label{eq:disc_kine}
z_i(\ell + 1)  = \left[ \begin{array}{c}
z_{i,1}(\ell+1) \\ z_{i,2}(\ell+1) \\ z_{i,3}(\ell+1) \end{array} \right] = \left[ \begin{array}{c} z_{i,1}(\ell) + g_1(\ell)\timestep \\  z_{i,2}(\ell) + g_2(\ell)\timestep \\ z_{i,3}(\ell) + g_3(\ell) \timestep \end{array} \right].
\end{align}
The binary sensor output is given by
{
\begin{align}\label{eq:output}
    h_i(z) = \begin{cases}
                 1 & \text{if } \exists j \neq i, s.t. ~z_j \in \operatorname{FOV}_i , \\
                 0 & \text{otherwise.}
             \end{cases}
\end{align}
}
where the FOV is the conical area in front of the sensor with a measured distance and opening angle chosen from the real robot distribution with average detection distance of 1.10m and FOV angle of 49 degrees. We then programmed the agents' sampling period to be the same as the real robots,~$\timestep=22$ms.

\bibliographystyle{ieeetr}
\bibliography{ricardo}

\begin{thebibliography}{10}

\bibitem{GZ-GKF-DPG:11}
G.~Zhang, G.~Fricke, and D.~Garg, ``Spill detection and perimeter surveillance via distributed swarming agents,'' {\em IEEE/asme Transactions on Mechatronics}, vol.~18, no.~1, pp.~121--129, 2011.

\bibitem{HK-CWF-IT-BS-ER-KP-AW-JW-12}
H.~Kuntze, C.~Frey, I.Tchouchenkov, B.~Staehle, E.~Rome, K.~Pfeiffer, A.~Wenzel, and J.~W{\"o}llenstein, ``Seneka-sensor network with mobile robots for disaster management,'' in {\em 2012 IEEE Conference on Technologies for Homeland Security (HST)}, pp.~406--410, IEEE, 2012.

\bibitem{MS-JC-LP-JT-GL-AT-VV-VK:14}
M.~Saska, J.~Chudoba, L.~P{\v{r}}eu{\v{c}}il, J.~Thomas, G.~Loianno, A.~T{\v{r}}e{\v{s}}{\v{n}}{\'a}k, V.~Von{\'a}sek, and V.~Kumar, ``Autonomous deployment of swarms of micro-aerial vehicles in cooperative surveillance,'' in {\em 2014 International Conference on Unmanned Aircraft Systems (ICUAS)}, pp.~584--595, IEEE, 2014.

\bibitem{RA-JJ-BA-EM:20}
R.~Arnold, J.~Jablonski, B.~Abruzzo, and E.~Mezzacappa, ``Heterogeneous uav multi-role swarming behaviors for search and rescue,'' in {\em 2020 IEEE Conference on Cognitive and Computational Aspects of Situation Management (CogSIMA)}, pp.~122--128, IEEE, 2020.

\bibitem{JT:05}
J.~Fromm, ``Types and forms of emergence,'' {\em arXiv preprint nlin/0506028}, 2005.

\bibitem{JT:05ten}
J.~Fromm, ``Ten questions about emergence,'' {\em arXiv preprint nlin/0509049}, 2005.

\bibitem{OTH:07}
O.~Holland, ``Taxonomy for the modeling and simulation of emergent behavior systems,'' in {\em Proceedings of the 2007 spring simulation multiconference-Volume 2}, pp.~28--35, 2007.

\bibitem{HH-TA-AR:20}
H.~Hamann, T.~Aust, and A.~Reina, ``Guerrilla performance analysis for robot swarms: Degrees of collaboration and chains of interference events,'' in {\em International Conference on Swarm Intelligence}, pp.~134--147, Springer, 2020.

\bibitem{GV-CV-GS-TN-AEE-TV:18}
G.~V{\'a}s{\'a}rhelyi, C.~Vir{\'a}gh, G.~Somorjai, T.~Nepusz, A.~Eiben, and T.~Vicsek, ``Optimized flocking of autonomous drones in confined environments,'' {\em Science Robotics}, vol.~3, no.~20, p.~eaat3536, 2018.

\bibitem{AO-MG-AK-MDH-RG:19}
A.~{\"O}zdemir, M.~Gauci, A.~Kolling, M.~Hall, and R.~Gro{\ss}, ``Spatial coverage without computation,'' in {\em 2019 International Conference on Robotics and Automation (ICRA)}, pp.~9674--9680, IEEE, 2019.

\bibitem{MG-JC-WL-TJD-RG:14}
M.~Gauci, J.~Chen, W.~Li, T.~Dodd, and R.~Gro{\ss}, ``Self-organized aggregation without computation,'' {\em The International Journal of Robotics Research}, vol.~33, no.~8, pp.~1145--1161, 2014.

\bibitem{DS-CP-GB:18}
D.~St-Onge, C.~Pinciroli, and G.~Beltrame, ``Circle formation with computation-free robots shows emergent behavioral structure,'' in {\em IEEE/RSJ International Conference on Intelligent Robots and Systems (IROS)}, pp.~5344--5349, IEEE, 2018.

\bibitem{MG-JC-WL-TJD-RG:14clust}
M.~Gauci, J.~Chen, W.~Li, T.~Dodd, and R.~Gro{\ss}, ``Clustering objects with robots that do not compute,'' in {\em Proceedings of the 2014 international conference on Autonomous agents and multi-agent systems}, pp.~421--428, 2014.

\bibitem{ALA-MGCAC-NDF-ML-GV:18}
A.~Alfeo, M.~Cimino, N.~D. Francesco, M.~Lega, and G.~Vaglini, ``Design and simulation of the emergent behavior of small drones swarming for distributed target localization,'' {\em Journal of computational science}, vol.~29, pp.~19--33, 2018.

\bibitem{YC-YRH-MRD-ALB:07}
Y.~Chuang, Y.~R. Huang, M.~R. D'Orsogna, and A.~L. Bertozzi, ``Multi-vehicle flocking: scalability of cooperative control algorithms using pairwise potentials,'' in {\em Proceedings 2007 IEEE international conference on robotics and automation}, pp.~2292--2299, IEEE, 2007.

\bibitem{AO-MG-RG:17}
A.~{\"O}zdemir, M.~Gauci, and R.~Gro{\ss}, ``Shepherding with robots that do not compute,'' in {\em Artificial Life Conference Proceedings}, pp.~332--339, MIT Press One Rogers Street, Cambridge, MA 02142-1209, USA journals-info~…, 2017.

\bibitem{YM-HG-YJ:13}
Y.~Meng, H.~Guo, and Y.~Jin, ``A morphogenetic approach to flexible and robust shape formation for swarm robotic systems,'' {\em Robotics and Autonomous Systems}, vol.~61, no.~1, pp.~25--38, 2013.

\bibitem{MR-AC-JW-GH-JM-RN:13}
M.~Rubenstein, A.~Cabrera, J.~Werfel, G.~Habibi, J.~McLurkin, and R.~Nagpal, ``Collective transport of complex objects by simple robots: theory and experiments,'' in {\em Proceedings of the 2013 international conference on Autonomous agents and multi-agent systems}, pp.~47--54, 2013.

\bibitem{HVDP-SB-JO:03}
H.~Parunak, S.~Brueckner, and J.~Odell, ``Swarming coordination of multiple uav's for collaborative sensing,'' in {\em 2nd AIAA" Unmanned Unlimited" Conf. and Workshop \& Exhibit}, p.~6525, 2003.

\bibitem{ALA-MGCAC-NDF-AL-ML-GV:18}
A.~L. Alfeo, M.~G. C.~A. Cimino, N.~D. Francesco, A.~Lazzeri, M.~Lega, and G.~Vaglini, ``Swarm coordination of mini-uavs for target search using imperfect sensors,'' {\em Intelligent Decision Technologies}, vol.~12, no.~2, pp.~149--162, 2018.

\bibitem{RAB:90}
R.~A. Brooks, ``Elephants don't play chess,'' {\em Robotics and autonomous systems}, vol.~6, no.~1-2, pp.~3--15, 1990.

\bibitem{RAB:91}
R.~A. Brooks, ``Intelligence without representation,'' {\em Artificial intelligence}, vol.~47, no.~1-3, pp.~139--159, 1991.

\bibitem{DW:06}
D.~Willshaw, ``Self-organization in the nervous system,'' {\em Cognitive systems: Information processing meet brain science}, pp.~5--33, 2006.

\bibitem{CT-CL-CN:20}
C.~Taylor, C.~Luzzi, and C.~Nowzari, ``On the effects of collision avoidance on emergent swarm behavior,'' in {\em 2020 American Control Conference (ACC)}, pp.~931--936, IEEE, 2020.

\bibitem{UB-LW-ADA-ME:15}
U.~Borrmann, L.~Wang, A.~D. Ames, and M.~Egerstedt, ``Control barrier certificates for safe swarm behavior,'' {\em IFAC-PapersOnLine}, vol.~48, no.~27, pp.~68--73, 2015.

\bibitem{KS-IBS-LMTR-CRH-DM-MAH:16}
K.~Szwaykowska, I.~B. Schwartz, L.~M. y~Teran~Romero, C.~R. Heckman, D.~Mox, and M.~A. Hsieh, ``Collective motion patterns of swarms with delay coupling: Theory and experiment,'' {\em Physical Review E}, vol.~93, no.~3, p.~032307, 2016.

\bibitem{HS:09}
H.~Sayama, ``Swarm chemistry,'' {\em Artificial life}, vol.~15, no.~1, pp.~105--114, 2009.

\bibitem{HS:11}
H.~Sayama, ``Seeking open-ended evolution in swarm chemistry,'' in {\em 2011 IEEE Symposium on Artificial Life (ALIFE)}, pp.~186--193, IEEE, 2011.

\bibitem{HS:12}
H.~Sayama, ``Morphologies of self-organizing swarms in 3d swarm chemistry,'' in {\em Proceedings of the 14th annual conference on Genetic and evolutionary computation}, pp.~577--584, 2012.

\bibitem{HS:07}
H.~Sayama, {\em Swarm Chemistry Homepage}.
\newblock Binghamton University https://bingdev.binghamton.edu/sayama/SwarmChemistry/, 2007.

\bibitem{VL-HH-LYC-JW-JI-DS-ML-KG:21}
V.~Lim, H.~Huang, L.~Y. Chen, J.~Wang, J.~Ichnowski, D.~Seita, M.~Laskey, and K.~Goldberg, ``Planar robot casting with real2sim2real self-supervised learning,'' {\em arXiv preprint arXiv:2111.04814}, 2021.

\bibitem{YC-AH-VM-MM-JI-NR-DF:19}
Y.~Chebotar, A.~Handa, V.~Makoviychuk, M.~Macklin, J.~Issac, N.~Ratliff, and D.~Fox, ``Closing the sim-to-real loop: Adapting simulation randomization with real world experience,'' in {\em 2019 International Conference on Robotics and Automation (ICRA)}, pp.~8973--8979, IEEE, 2019.

\bibitem{LW-RG-QV-YQ-HS-HC:22}
L.~Wang, R.~Guo, Q.~Vuong, Y.~Qin, H.~Su, and H.~Christensen, ``A real2sim2real method for robust object grasping with neural surface reconstruction,'' {\em arXiv preprint arXiv:2210.02685}, 2022.

\bibitem{OP-GE-AJC:09}
O.~Paunovski, G.~Eleftherakis, and A.~Cowling, ``Disciplined exploration of emergence using multi-agent simulation framework,'' {\em Computing and Informatics}, vol.~28, no.~3, pp.~369--391, 2009.

\bibitem{VD:94}
V.~Darley, ``Emergent phenomena and complexity,'' {\em Artificial Life}, vol.~4, pp.~411--416, 1994.

\bibitem{YBY:19}
Y.~Bar-Yam, {\em Dynamics of complex systems}.
\newblock CRC Press, 2019.

\bibitem{RC-CC:96}
R.~Conte and C.~Castelfranchi, ``Simulating multi-agent interdependencies. a two-way approach to the micro-macro link,'' in {\em Social science microsimulation}, pp.~394--415, Springer, 1996.

\bibitem{BE:05}
B.~Edmonds, ``Using the experimental method to produce reliable self-organised systems,'' {\em Engineering Self-Organising Systems}, vol.~3464, pp.~84--99, 2005.

\bibitem{JHH:00}
J.~Holland, {\em Emergence: From chaos to order}.
\newblock OUP Oxford, 2000.

\bibitem{HS-HKG:22}
H.~Sinhmar and H.~Kress-Gazit, ``Decentralized control of minimalistic robotic swarms for guaranteed target encapsulation,'' in {\em 2022 IEEE/RSJ International Conference on Intelligent Robots and Systems (IROS)}, pp.~7251--7258, IEEE, 2022.

\bibitem{ZC-ZC-VT-DC-HZ:16}
Z.~Cheng, Z.~Chen, T.~Vicsek, D.~Chen, and H.~Zhang, ``Pattern phase transitions of self-propelled particles: gases, crystals, liquids, and mills,'' {\em New Journal of Physics}, vol.~18, no.~10, p.~103005, 2016.

\bibitem{MRD-YC-ALB-LSC:06}
M.~R. D’Orsogna, Y.~Chuang, A.~L. Bertozzi, and L.~S. Chayes, ``Self-propelled particles with soft-core interactions: patterns, stability, and collapse,'' {\em Physical review letters}, vol.~96, no.~10, p.~104302, 2006.

\bibitem{NVB-HA-IYT-SAM:20}
N.~V. Brilliantov1, H.~Abutuqayqah, I.~Y. Tyukin, and S.~A. Matveev, ``Swirlonic state of active matter,'' {\em Nature Research Scientific Reports}, vol.~10, no.~16783, 2020.

\bibitem{RGS:10}
R.~Sargent, ``Verification and validation of simulation models,'' in {\em Proceedings of the 2010 winter simulation conference}, pp.~166--183, IEEE, 2010.

\bibitem{VB:96}
V.~Braitenberg, {\em Vehicles: Experiments in synthetic psychology}.
\newblock MIT press, 1986.

\bibitem{MG-JC-TJD-RG:14}
M.~Gauci, J.~Chen, T.~J. Dodd, and R.~Gro{\ss}, ``Evolving aggregation behaviors in multi-robot systems with binary sensors,'' in {\em Distributed Autonomous Robotic Systems} (M.~{Ani Hsieh} and G.~Chirikjian, eds.), (Berlin, Heidelberg), pp.~355--367, Springer Berlin Heidelberg, 2014.

\bibitem{DB-RT-OH-SL:18}
D.~S. Brown, R.~Turner, O.~Hennigh, and S.~Loscalzo, ``Discovery and exploration of novel swarm behaviors given limited robot capabilities,'' in {\em Distributed Autonomous Robotic Systems}, Springer International Publishing AG, 2018.

\bibitem{FB-MG-RN:21}
F.~Berlinger, M.~Gauci, and R.~Nagpal, ``Implicit coordination for 3d underwater collective behaviors in a fish-inspired robot swarm,'' {\em Science Robotics}, vol.~6, no.~50, 2021.

\bibitem{CN-JC:11-auto}
C.~Nowzari and J.~Cort\'es, ``Self-triggered coordination of robotic networks for optimal deployment,'' {\em Automatica}, vol.~48, no.~6, pp.~1077--1087, 2012.

\bibitem{JC-SM-TK-FB:02-tra}
J.~Cort{\'e}s, S.~Mart{\'\i}nez, T.~Karatas, and F.~Bullo, ``Coverage control for mobile sensing networks,'' {\em IEEE Transactions on robotics and Automation}, vol.~20, no.~2, pp.~243--255, 2004.

\bibitem{AC-CKH:18}
A.~Costanzo and C.~K. Hemelrijk, ``Spontaneous emergence of milling (vortex state) in a vicsek-like model,'' {\em Journal of Physics D: Applied Physics}, vol.~51, no.~13, p.~134004, 2018.

\bibitem{UW:99}
U.~Wilensky, {\em NetLogo (and NetLogo User Manual)}.
\newblock Center for Connected Learning and Computer-Based Modeling, Northwestern University. http://ccl.northwestern.edu/netlogo/, 1999.

\end{thebibliography}

\end{document}